\documentclass[conference]{IEEEtran}
\IEEEoverridecommandlockouts
\usepackage{cite}
\usepackage{amsmath,amssymb,amsfonts}
\usepackage{graphicx}
\usepackage{textcomp}
\usepackage{xcolor}
\usepackage{booktabs}
\usepackage{url}
\usepackage{balance}
\usepackage{caption}

\def\BibTeX{{\rm B\kern-.05em{\sc i\kern-.025em b}\kern-.08em
    T\kern-.1667em\lower.7ex\hbox{E}\kern-.125emX}}
\begin{document}

\newcommand{\model}{StructAttn}

\title{Structured-Token Attention for Fine-Grained\\mmWave Liquid Identification\\
}

\author{\IEEEauthorblockN{Junseo Min}
\IEEEauthorblockA{\textit{Department of Defense Convergence Technology} \\
\textit{Korea University}\\
Seoul, Republic of Korea \\
junseo\_min@korea.ac.kr}
\and
\IEEEauthorblockN{Joongheon Kim}
\IEEEauthorblockA{\textit{Department of Electrical \& Computer Engineering} \\
\textit{Korea University}\\
Seoul, Republic of Korea \\
joongheon@korea.ac.kr}
}

\maketitle

\setlength{\abovedisplayskip}{6pt plus 2pt minus 2pt}
\setlength{\belowdisplayskip}{6pt plus 2pt minus 2pt}
\setlength{\abovedisplayshortskip}{3pt plus 1pt minus 1pt}
\setlength{\belowdisplayshortskip}{3pt plus 1pt minus 1pt}

\begin{abstract}
Fine-grained liquid identification requires a classifier to distinguish similar substances from compact millimeter-wave measurements. We investigate structured-token attention, which represents a radar feature tensor through complementary antenna, feature, branch, and global views. The proposed StructAttn model embeds these views into eleven tokens and combines them with a three-layer transformer. Experiments on the public FG-LiquID data cover 30 liquids and 3,000 samples. Over five seeds on the released split, StructAttn achieves 96.70\% mean best-epoch accuracy, matching the original-recipe MMNet baseline and exceeding a scalar-token attention model by 1.50 percentage points. A five-seed probability ensemble reaches 97\%, corresponding to 18 errors among 600 test samples. Final-epoch and five-fold results complement the checkpoint-selected comparison, with StructAttn obtaining 95.30\% out-of-fold accuracy. Token and feature analyses show that structured views are useful for prediction, while residual confusions concentrate in water, soft-drink, and edible-oil classes. These results support structured tokenization as a practical design choice for attention-based liquid recognition.
\end{abstract}

\section{Introduction}
Contactless liquid identification is relevant to automated inspection and everyday sensing. Similar-looking liquids can belong to different categories, and different containers can hold the same substance. Millimeter-wave (mmWave) measurements provide an alternative source of information through the interaction between electromagnetic signals and the observed liquid. FG-LiquID demonstrates fine-grained recognition using a learned representation of radar-derived features~\cite{fgliquid}.

A compact sensing tensor contains several kinds of structure. Measurements from one receiving antenna describe a common sensing channel, whereas values of the same feature across antennas describe a common observable. Location-related features and reflected signal strength also offer different views of a sample. Flattening this tensor into independent scalars leaves these relationships implicit and requires the classifier to recover them from limited training data.

Self-attention provides a mechanism for combining measurements according to the current sample~\cite{transformer}. Its usefulness depends on token construction. Unlike a scalar token, a complete antenna or feature slice encodes local relationships before interacting with other tokens. This motivates structured input views.

This paper develops \model, a structured-token attention model for fine-grained mmWave liquid identification. The model constructs overlapping antenna, feature, location, reflection, and global views without adding measurements. A small token sequence is processed by a transformer, and the output combines a learned classification token with the mean of the measurement tokens. The design retains the semantics of the input dimensions while allowing interactions across views.

The evaluation examines structured tokenization against attention-based alternatives, compares MMNet under different checkpoint and aggregation settings, and identifies informative measurements and confusable liquids. It uses repeated seeds, token variants, feature masking, and out-of-fold predictions.

The main contribution is a measurement-aware token construction and its empirical characterization on FG-LiquID. A single \model-S3 model reaches the same mean best-epoch accuracy as original-recipe MMNet, while a five-seed ensemble attains the highest ensemble accuracy among the compared configurations. Best-epoch, final-epoch, and cross-validation results are interpreted separately.

\begin{figure*}[t]
\centering
\vspace{-10mm}
\includegraphics[width=\textwidth]{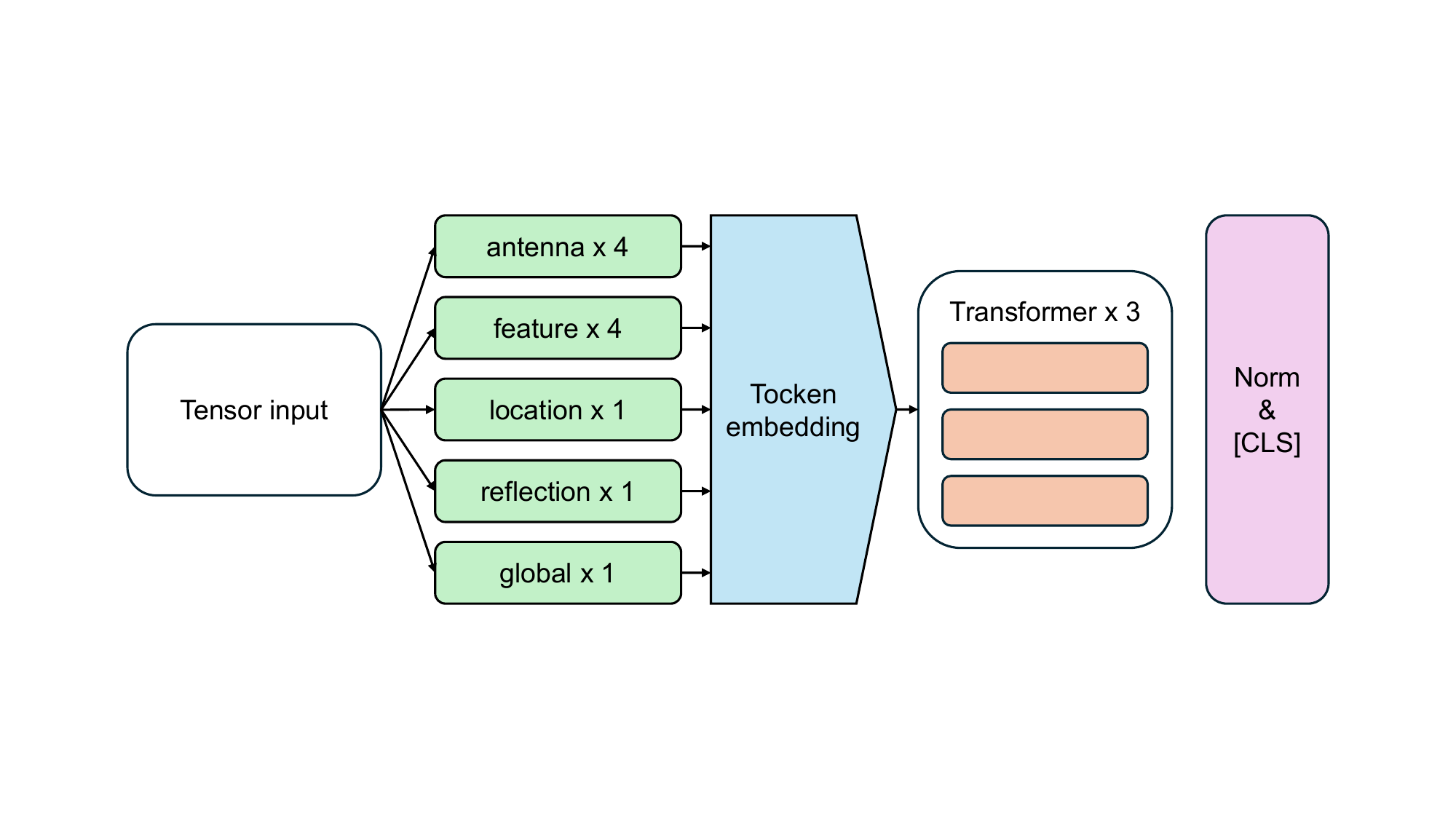}
\vspace{-30mm}
\caption{Structured views and attention processing in StructAttn-S3.}
\vspace{-5mm}
\label{fig:architecture}
\end{figure*}

\section{Related Work}
\subsection{Wireless Liquid Identification}
mmWave studies examine 60 GHz connectivity~\cite{kim2017fog} and antenna-array link budgets~\cite{kim2017maa}. Related work addresses receive beamforming~\cite{kim2014beam}, two-stage beam training~\cite{kim2015beam}, and low-power link setup~\cite{kim2014link}. Frequency-sharing analysis~\cite{han2020sharing} and joint clustering, beamforming, and power allocation~\cite{lim2024noma} characterize interference in communication systems. These studies provide physical and system context for mmWave measurements, while liquid recognition requires a task-specific classifier.

FG-LiquID uses mmWave sensing for fine-grained liquid recognition and introduces MMNet to process complementary measurement features~\cite{fgliquid}. Its feature representation is used in our evaluation. We retain the distinction between location-related measurements and reflection strength, but expose additional antenna and feature views to a shared attention module.

Wireless liquid sensing also includes approaches based on WiFi. LiquImager jointly addresses liquid identification and container imaging using commercial WiFi devices~\cite{liquimager}. This line of work emphasizes the sensing configuration and container-related information. Our focus is instead the classifier applied to an existing mmWave feature tensor. Cross-platform recognition rates are therefore not comparable.

\subsection{Attention for Structured Measurements}
Wireless inference also depends on the task assigned to received measurements. Choi and Kim analyze blind signal classification errors for interference-cancellation decisions in nonorthogonal multiple access~\cite{choi2020blind}. Lim et al. use deep residual learning for pilot design and channel estimation under hardware impairments~\cite{lim2022pilot}. Our task instead classifies liquids from an already extracted radar feature tensor, with structured tokens organizing the available observables.

The transformer models interactions using query, key, and value projections~\cite{transformer}. Set Transformer extends attention-based processing to sets of instances~\cite{settransformer}. These methods motivate relational aggregation, although our tokens are not interchangeable set elements. Antenna identity, feature identity, and token type are preserved through learned embeddings.

FT-Transformer adapts transformer processing to tabular features and highlights the importance of strong neural baselines under consistent evaluation protocols~\cite{fttransformer}. Its feature-token perspective is relevant to compact sensor inputs. \model\ extends this perspective to overlapping multi-value views of a three-dimensional measurement tensor. AttnNet is a scalar-token baseline, not a reproduction of FT-Transformer. Multiscale LSTM forecasting similarly combines representations at different temporal scales~\cite{kim2021multiscale}, whereas our views follow antenna and feature dimensions.

Adaptive perception studies balance accuracy and processing cost through SAR filter selection~\cite{kim2022sar}, optical-flow-assisted detection~\cite{yun2023detection}, and joint frame-rate and recognition-model selection~\cite{kim2023uav}. These system-level approaches complement our focus on representation design for fixed radar features.

\subsection{Training and Prediction Aggregation}
Optimizer and regularization choices can influence the apparent benefit of an architecture. AdamW separates weight decay from adaptive gradient updates~\cite{adamw}. We therefore include MMNet-R, which keeps the MMNet architecture and applies the same modern training recipe used by the attention models. This comparison helps distinguish architecture changes from training-recipe effects.

Combining independently trained networks is another way to aggregate predictions~\cite{ensemble}. We evaluate probability averaging over five seeds for MMNet, MMNet-R, and \model-S3. Ensemble results are reported separately from single-model means because they describe different predictors and require multiple forward passes. The study evaluates classification performance without making an uncertainty-calibration claim.

\section{Method}
\subsection{Task and Input Representation}
Let $X\in\mathbb{R}^{P\times F\times R}$ denote one liquid measurement, where $P=3$ data points, $F=4$ feature types, and $R=4$ receiving antennas. The feature types are distance, elevation angle of arrival, azimuth angle of arrival, and received signal strength (RSS). The classifier estimates a label $y\in\{1,\ldots,C\}$ with $C=30$. Each sample therefore contains 48 scalar values.

For the modern training recipe, each input coordinate is standardized using training-set statistics. With coordinate index $j$, training mean $\mu_j$, and training standard deviation $\sigma_j$, the normalized value is
\begin{equation}
 \widetilde{x}_j=\frac{x_j-\mu_j}{\max(\sigma_j,\epsilon)},
\end{equation}
where $\epsilon>0$ prevents division by zero. Test measurements use the same training statistics. This preserves coordinate identities while aligning numerical scales.

\subsection{Structured Token Construction}
Figure~\ref{fig:architecture} summarizes the model. Each token is obtained by extracting and vectorizing a predefined slice of $\widetilde{X}$. Antenna tokens collect all data points and feature types for a fixed receiver. Feature tokens collect one observable across data points and receivers. These two views make the channel-wise and observable-wise organization available before attention is applied.

The location token concatenates distance and the two angle features, while the reflection token contains RSS. A global token includes the entire tensor. The reflection token and the RSS feature token contain the same input coordinates, but occupy distinct semantic roles in the token sequence. Their identities allow the network to use this information in both the feature and branch representations.

\begin{table}[t]
\caption{Token construction for StructAttn-S3.}
\label{tab:tokens}
\centering\small
\begin{tabular}{lccp{3.0cm}}
\toprule
View & Count & Values & Grouping\\
\midrule
Antenna & 4 & 12 & One receiver\\
Feature & 4 & 12 & One observable\\
Location & 1 & 36 & Distance, angles\\
Reflection & 1 & 12 & RSS branch\\
Global & 1 & 48 & Complete tensor\\
\midrule
Total & 11 & -- & Without [CLS]\\
\bottomrule
\vspace{-8mm}
\end{tabular}
\end{table}

The overlapping groups in Table~\ref{tab:tokens} summarize the same measurement without adding observations. They provide structured views for learning interactions. A separate data-point view is examined in the token-variant experiments.

\subsection{Embedding and Attention Encoder}
For token $t$, let $g_t(\widetilde{X})\in\mathbb{R}^{m_t}$ denote its extracted sub-vector. A token embedder projects it to width $d=192$. Each embedder follows a linear, batch-normalization, GELU, dropout, and linear sequence with hidden width 256. The embedded token is written as
\begin{equation}
 z_t=\phi_t\bigl(g_t(\widetilde{X})\bigr)+e_t+e_{\tau(t)},
\end{equation}
where $e_t$ encodes token identity and $e_{\tau(t)}$ encodes token type. The notation $\phi_t$ denotes the type-specific embedder assigned to token $t$.

A learned classification vector $z_{\mathrm{cls}}$ is prepended to the eleven measurement tokens. The initial sequence $H^{(0)}\in\mathbb{R}^{12\times192}$ passes through three pre-layer-normalized transformer layers. Each layer uses six attention heads, a feed-forward hidden width of 384, and dropout of 0.1.

For one attention head, the normalized sequence $U$ is mapped to query, key, and value matrices. The operation is
\begin{equation}
 Q=UW_Q,\quad K=UW_K,\quad V=UW_V,
\end{equation}
\begin{equation}
 \operatorname{Attn}(U)=\operatorname{softmax}\left(\frac{QK^\top}{\sqrt{d_h}}\right)V,
\end{equation}
where $d_h=32$. Outputs of the six heads are concatenated and projected. A pre-normalized layer then applies residual updates
\begin{align}
 \widehat{H}^{(\ell)}&=H^{(\ell-1)}+\operatorname{MHA}\bigl(\operatorname{LN}(H^{(\ell-1)})\bigr),\\
 H^{(\ell)}&=\widehat{H}^{(\ell)}+\operatorname{FFN}\bigl(\operatorname{LN}(\widehat{H}^{(\ell)})\bigr).
\end{align}
Dropout is applied within these blocks. The attention weights vary with the input and therefore allow different samples to combine the structured views differently.

\subsection{Readout and Learning Objective}
The classifier combines the final classification token with the average of the eleven measurement-token outputs. With $L=3$ and $T=11$, the representation is
\begin{equation}
 r=\operatorname{LN}\left(h_{\mathrm{cls}}^{(L)}\oplus\frac{1}{T}\sum_{t=1}^{T}h_t^{(L)}\right),
\end{equation}
where $\oplus$ denotes concatenation. A linear classifier followed by softmax produces $p_\theta(c\mid X)$. The readout combines learned and distributed summaries.

Training uses cross-entropy with label smoothing. For smoothing coefficient $\alpha=0.1$, the target distribution and loss are
\begin{equation}
 \begin{aligned}
 q_c&=(1-\alpha)\mathbb{1}[c=y]+\frac{\alpha}{C},\\
 \mathcal{L}&=-\sum_{c=1}^{C}q_c\log p_\theta(c\mid X).
 \end{aligned}
\end{equation}
Gaussian input noise with standard deviation 0.05 is used in the modern recipe. The classifier predicts the class with maximum output probability.

\subsection{Ensemble and Model Size}
The five-seed ensemble averages probabilities from independently trained checkpoints. Its prediction is
\begin{equation}
 \widehat{y}_{\mathrm{ens}}=\arg\max_c\frac{1}{5}\sum_{s=1}^{5}p_{\theta_s}(c\mid X).
\end{equation}
Aggregation requires no additional classifier and retains the input representation and label space.

\model-S3 contains 1,187,742 trainable parameters. Attention uses twelve positions including [CLS], giving 144 pairs per head. Embedding and projection layers also contribute to cost. The model is larger than the 238K-parameter MMNet baseline, and the present comparison does not assume lower inference latency or energy consumption.

\begin{table*}[t]
\caption{Released-split performance over five seeds. Accuracy and macro-F1 are percentages.}
\label{tab:main}
\centering\small
\begin{tabular}{lccccc}
\toprule
Model & Parameters & Best epoch & Final epoch & Macro-F1 & Five-seed ensemble\\
\midrule
Nearest centroid & 0 &70.17&70.17&69.31&--\\
MLP &50K&$94.47\pm0.22$&$93.73\pm0.45$&94.23&--\\
MMNet &238K&$96.70\pm0.70$&$\mathbf{96.30\pm0.76}$&96.66&96.33 (22 errors)\\
MMNet-R &238K&$\mathbf{96.93\pm0.09}$&$96.23\pm0.19$&\textbf{96.85}&96.67 (20 errors)\\
NoAttn &283K&$92.37\pm0.14$&$91.43\pm0.25$&92.10&--\\
AttnNet &548K&$95.20\pm0.76$&$94.53\pm0.92$&94.84&--\\
\textbf{StructAttn-S3} &1.19M&$96.70\pm0.32$&$95.93\pm0.49$&96.56&\textbf{97.00 (18 errors)}\\
StructAttn-S2 &2.56M&$96.70\pm0.18$&$95.77\pm0.40$&96.57&--\\
StructAttn-S5 &1.19M&$96.43\pm0.37$&$95.90\pm0.43$&96.28&--\\
StructAttn-S6 &1.19M&$94.17\pm0.29$&$92.97\pm0.38$&93.83&--\\
\bottomrule
\vspace{-8mm}
\end{tabular}
\end{table*}

\begin{table}[t]
\caption{Composition of the released FG-LiquID split.}
\label{tab:data}
\centering\small
\begin{tabular}{lrrr}
\toprule
Category & Classes & Train & Test\\\midrule
Liquor &12&958&242\\
Edible oil &5&393&107\\
Soft drink &5&409&91\\
Water &2&161&39\\
Milk &2&157&43\\
Beer and wine &2&162&38\\
Coffee &1&78&22\\
Vinegar &1&82&18\\\midrule
Total &30&2,400&600\\\bottomrule
\end{tabular}
\end{table}

\section{Experiments}
\subsection{Dataset and Evaluation Protocol}
The public FG-LiquID data contain 3,000 measurements from 30 liquids, with 100 samples per class. The released split contains 2,400 training and 600 test samples. Table~\ref{tab:data} summarizes the categories. The liquor classes span 22\%, 33\%, 38\%, 43\%, 52\%, 53\%, 55\%, 56\%, 60\%, 62\%, 68\%, and 75\% alcohol by volume. These classes represent liquor labels rather than a synthetic ethanol--water mixture experiment.

Five training seeds are used for the main single-model comparison. Best-epoch accuracy is the maximum test accuracy across the training trajectory, and the reported mean and sample standard deviation summarize the five runs. Because the release has no dedicated validation split, these checkpoints were selected using the test set. The S3 configuration was also chosen using seed-0 test accuracy. Best-epoch and ensemble scores are therefore checkpoint-selected comparisons rather than estimates from an untouched model-selection holdout.

Final-epoch accuracy is reported alongside the best-epoch results to remove epoch selection from that comparison. A further evaluation uses stratified five-fold cross-validation over all 3,000 samples with seed 0 and final-epoch predictions. Each sample contributes one out-of-fold label, evaluating sample-level resampling rather than independent acquisition sessions.

Accuracy measures the fraction of correct predictions. Macro-F1 averages the class-wise F1 values over all 30 labels, providing a complementary measure when the released test counts vary by class. A single error changes overall test accuracy by approximately 0.167 percentage points. Class-level test counts range from 12 to 29, so individual class percentages are interpreted together with their sample counts.

\subsection{Baselines and Training Settings}
The comparison includes a nearest-centroid reference, a multilayer perceptron (MLP), original-recipe MMNet, and MMNet-R. Nearest centroid provides a feature-space reference without gradient training. NoAttn uses a token model without attention, whereas AttnNet uses per-scalar tokens. The baselines span direct feature classification and token interaction.

The modern recipe uses AdamW with learning rate $10^{-3}$, weight decay 0.05, batch size 128, and 300 epochs. The schedule uses ten warm-up epochs followed by cosine decay. Training-statistic normalization, label smoothing of 0.1, and Gaussian input noise of 0.05 are applied to MMNet-R and the attention models. Thus, MMNet-R is a training-recipe control for the proposed architecture.

Original-recipe MMNet uses Adam with learning rate 0.005, momentum coefficients 0.9 and 0.99, batch size 480, and 300 epochs. Its step schedule multiplies the learning rate by 0.8 every ten epochs. Weight decay of $10^{-5}$ is combined with explicit $L_2$ regularization of $9\times10^{-5}$. The MLP uses 200 epochs, batch size 128, learning rate 0.001, weight decay $10^{-4}$, dropout 0.3, and a factor-0.5 learning-rate step every 40 epochs.

\subsection{Main Recognition Results}
Table~\ref{tab:main} reports the main results. \model-S3 achieves 96.70\% mean best-epoch accuracy, matching MMNet and improving on AttnNet by 1.50 percentage points. The NoAttn-to-AttnNet difference is 2.83 points. These comparisons support token interaction and structured views within the tested model family, although parameter counts differ and the comparisons do not isolate attention under equal capacity.

\begin{figure}[t]
\centering\includegraphics[width=0.85\columnwidth]{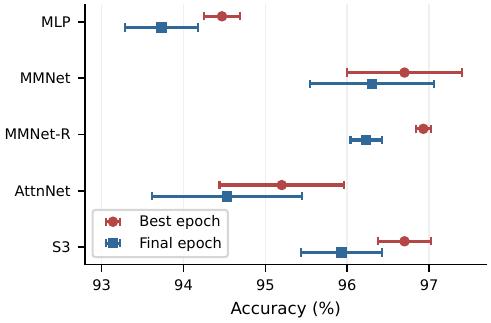}
\caption{Best- and final-epoch accuracy over five seeds.}
\label{fig:performance}
\vspace{-3mm}
\end{figure}

MMNet-R obtains the highest single-model best-epoch mean of 96.93\%. The original-recipe MMNet gives the highest final-epoch mean of 96.30\%, compared with 95.93\% for S3. The best-to-final gap is 0.77 points for S3 and 0.70 points for MMNet-R. Figure~\ref{fig:performance} shows these protocol-dependent differences.

The S3 ensemble reaches 97.00\%, with 18 errors among 600 samples. The MMNet and MMNet-R ensembles make 22 and 20 errors, respectively. Thus, S3 improves the ensemble result by four and two correct predictions. This small margin is the best observed ensemble score, without a statistical-significance claim.

The MMNet and MMNet-R ensemble members were retrained separately from the main seed-sweep runs. Their mean selected-checkpoint accuracies are 96.57\% and 96.87\%, respectively, while the S3 member mean is 96.70\%. Ensemble accuracy is therefore compared directly between ensemble predictors, rather than inferred from the single-model means in Table~\ref{tab:main}.

\subsection{Token Design and Regularization}
Table~\ref{tab:variants} compares token variants using seed 0. S1 contains antenna, feature, data-point, branch, and global tokens. S3 removes the three data-point tokens while keeping width 192 and three layers. Its best-epoch accuracy is 97.00\%, compared with 96.33\% for S1. This screening supports retaining antenna and feature views without a separate token for each data point.

\begin{table}[t]
\caption{Token variants on seed 0. Accuracy is in percent.}
\label{tab:variants}
\centering\small
\begin{tabular}{lrrrrr}
\toprule
Variant & Tokens & Width/layers & Params & Best & Final\\\midrule
S1 &14&192/3&1.24M&96.33&96.00\\
S2 &14&256/4&2.56M&96.83&96.00\\
S3 &11&192/3&1.19M&97.00&96.50\\
S4 &3&192/2&782K&96.33&95.00\\\bottomrule
\vspace{-5mm}
\end{tabular}
\end{table}

S4 retains only the location, reflection, and global tokens and uses two transformer layers. It reaches 95.00\% final-epoch accuracy, compared with 96.50\% for S3 in this screening. The comparison is consistent with a benefit from richer structured views, but token removal and reduced depth occur together. This comparison does not isolate each token group's contribution.

The larger S2 model does not improve the five-seed best-epoch mean over S3, despite using approximately twice as many parameters. S5 adds an exponential moving average with decay 0.998 and obtains 96.43\%. S6 additionally uses dropout 0.2, mixup coefficient 0.2, and input noise of 0.1, reaching 94.17\%. The tested data favor the S3 configuration over these larger or more strongly regularized alternatives.

\subsection{Measurement Contributions}
\begin{figure}[t]
\centering\includegraphics[width=0.85\columnwidth]{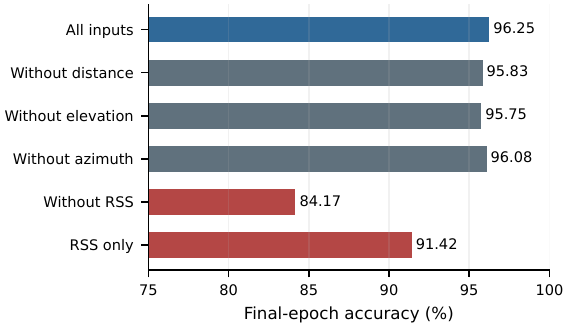}
\caption{Feature masking with MMNet-R at the final epoch.}
\label{fig:features}
\vspace{-3mm}
\end{figure}

Feature masking is evaluated by retraining MMNet-R with selected inputs removed or retained. The results use final-epoch accuracy averaged over two seeds. As shown in Fig.~\ref{fig:features}, removing RSS reduces accuracy from 96.25\% to 84.17\%. Keeping only RSS reaches 91.42\%. Reflection strength is therefore highly informative, while the other features provide complementary information.

Removing distance, elevation, or azimuth individually yields 95.83\%, 95.75\%, and 96.08\%, respectively. Removing a single receiving antenna gives accuracies between 96.08\% and 96.25\%. Using only the first data point reaches 95.50\%. These observations indicate substantial redundancy in the recorded feature tensor, which helps explain why additional data-point tokens need not improve recognition.

The recorded final-layer [CLS] attention patterns also vary by liquid category. Soft drinks emphasize the RSS feature token, oils and water emphasize distance, and liquor classes emphasize the reflection branch. These patterns describe model behavior rather than causal importance. Masking provides separate evidence about input usefulness.

\subsection{Cross-Validation and Class-Level Behavior}
Table~\ref{tab:cv} presents five-fold results. MMNet-R obtains 95.77\% out-of-fold accuracy and S3 obtains 95.30\%, with fold standard deviations of 0.40 and 0.64 points. This ranking differs from the ensemble comparison on the released split. The results indicate that structured attention is competitive on the available measurements, while its strongest observed aggregate score is specific to the tested ensemble setting.

\begin{table}[t]
\caption{Final-epoch five-fold results on all 3,000 samples.}
\label{tab:cv}
\centering\small
\begin{tabular}{lcc}
\toprule
Metric or category & MMNet-R & StructAttn-S3\\\midrule
Fold mean $\pm$ std & $95.77\pm0.40$ & $95.30\pm0.64$\\
Liquor &99.8&99.9\\
Edible oil &90.6&88.6\\
Soft drink &92.0&91.8\\
Water &84.5&82.5\\
Other categories &98.8&98.8\\\bottomrule
\vspace{-5mm}
\end{tabular}
\end{table}

Table~\ref{tab:confusion} gives representative S3 out-of-fold confusion counts. Liquor classes are separated almost perfectly in both models, whereas water, oil, and soft-drink categories account for most residual errors. For S3, mineral water has 74 correct predictions out of 100, with 14 assigned to tap water and 12 to soda. Olive oil has 81 correct predictions, with the remaining 19 assigned to soybean oil. Soybean oil has 83 correct predictions, with 17 assigned to olive oil. These bidirectional confusions are consistent with measured signatures.

The released-split per-class results provide a complementary view of ensemble behavior. Compared with one original MMNet run, the S3 ensemble reduces soda errors from five to one and corrects the two errors in each of the 38\% and 56\% liquor classes. Some classes move in the opposite direction, including mineral water. This comparison is diagnostic because it contrasts one run with an ensemble and uses only 12--29 samples per class.

\begin{table}[t]
\caption{Representative S3 out-of-fold confusions.}
\label{tab:confusion}
\centering\small
\begin{tabular}{lrlr}
\toprule
True liquid & Correct/100 & Predicted instead & Count\\\midrule
Mineral water &74&Tap water&14\\
 & &Soda&12\\
Olive oil &81&Soybean oil&19\\
Soybean oil &83&Olive oil&17\\
Soda &84&Tap water&10\\
 & &Mineral water&6\\
Corn oil &90&Peanut oil&9\\
Peanut oil &91&Corn oil&9\\\bottomrule
\end{tabular}
\end{table}

\subsection{Effect of the Number of Liquid Classes}
\begin{figure}[t]
\centering\includegraphics[width=0.85\columnwidth]{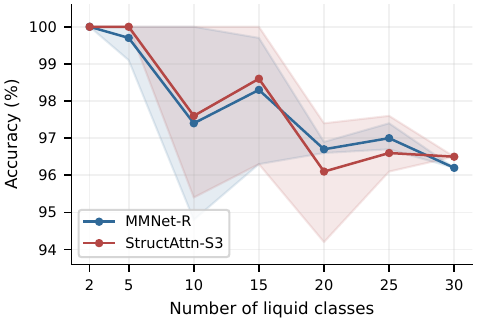}
\caption{Accuracy across sampled liquid subsets. Bands show ranges.}
\label{fig:classes}
\vspace{-3mm}
\end{figure}

The class-subset experiment restricts the released split to $K$ randomly selected liquids. Three subsets are evaluated for each $K<30$ using final-epoch predictions. Figure~\ref{fig:classes} shows the mean and observed range across subsets. Both models achieve 100\% for the sampled two-class tasks, and S3 also achieves 100\% for the sampled five-class tasks.

\balance
At $K=10$, S3 ranges from 95.4\% to 100\%, while at $K=15$ it ranges from 96.3\% to 100\%. Accuracy is therefore not determined by class count alone. Including a confusable pair can make a smaller task harder than a larger task composed of more separable liquids. Three sampled subsets characterize this dependence without covering all combinations.

Each liquid contributes 100 measurements, and subsets retain the released split. The ranges reflect label combinations rather than repeated training seeds. 

The $K=30$ endpoint is a single full-class run, with 96.2\% for MMNet-R and 96.5\% for S3 at the precision reported in the subset study. It is distinct from both the five-seed final-epoch means and the cross-validation results. Separate reporting clarifies the effect of class composition.

\section{Conclusion}
This paper presented structured-token attention for fine-grained mmWave liquid identification. Antenna, feature, branch, and global views expose complementary relationships in a compact radar tensor. On FG-LiquID, StructAttn-S3 matches the original-recipe MMNet best-epoch mean of 96.70\% and improves on scalar-token attention by 1.50 percentage points. Its five-seed ensemble reaches 97.00\%, while final-epoch and cross-validation results provide complementary views of performance.

The analysis identifies RSS as a central recognition cue and locates the remaining ambiguity mainly within similar liquid groups. Evaluation covers released measurements and sample-level splits. An independent validation partition and additional acquisition conditions are natural extensions for assessing configuration selection and transfer. Structured tokenization provides a basis for further attention-based liquid sensing.

\bibliographystyle{IEEEtran}
\bibliography{references}

@article{fgliquid,
author = {Liang, Yumeng and others},
title = {FG-LiquID: A Contact-less Fine-grained Liquid Identifier by Pushing the Limits of Millimeter-wave Sensing},
year = {2021},
issue_date = {Sept 2021},
publisher = {Association for Computing Machinery},
address = {New York, NY, USA},
volume = {5},
number = {3},
journal = {Proc. ACM Interact. Mob. Wearable Ubiquitous Technol.},
month = sep,
articleno = {116},
numpages = {27}
}

@inproceedings{transformer,
  author = {Vaswani, Ashish and others},
  title = {{Attention is All you Need}},
  booktitle = {Proc. Advances in Neural Information Processing Systems (NIPS)},
  address = {Long Beach, CA, USA},
  month = {December},
  year = {2017},
  pages = {5998--6008}
}

@article{kim2017fog,
  author = {Kim, Joongheon and Lee, Wonjun},
  title = {{Feasibility Study of 60 GHz Millimeter-Wave Technologies for Hyperconnected Fog Computing Applications}},
  journal = {IEEE Internet of Things Journal},
  volume = {4},
  number = {5},
  pages = {1165--1173},
  month = {October},
  year = {2017}
}

@article{kim2017maa,
  author = {Kim, Joongheon and others},
  title = {{60 GHz Modular Antenna Array Link Budget Estimation with WiGig Baseband and Millimeter-Wave Specific Attenuation}},
  journal = {International Journal of Antennas and Propagation},
  volume = {2017},
  pages = {1--9},
  month = {June},
  year = {2017}
}

@article{liquimager,
  author = {Shang, Fei and others},
  title = {{LiquImager: Fine-grained Liquid Identification and Container Imaging System with COTS WiFi Devices}},
  journal = {Proc. of the ACM on Interactive, Mobile, Wearable and Ubiquitous Technologies},
  volume = {8},
  number = {1},
  pages = {1--29},
  month = {March},
  year = {2024}
}

@article{choi2020blind,
  author = {Choi, Minseok and Kim, Joongheon},
  title = {{Blind Signal Classification Analysis and Impact on User Pairing and Power Allocation in Nonorthogonal Multiple Access}},
  journal = {IEEE Access},
  volume = {8},
  pages = {100916--100929},
  month = {June},
  year = {2020}
}

@article{lim2022pilot,
  author = {Lim, Byungju and others},
  title = {{Joint Pilot Design and Channel Estimation Using Deep Residual Learning for Multi-Cell Massive MIMO Under Hardware Impairments}},
  journal = {IEEE Transactions on Vehicular Technology},
  volume = {71},
  number = {7},
  pages = {7599--7612},
  month = {July},
  year = {2022}
}

@inproceedings{settransformer,
  author = {Lee, Juho and others},
  title = {{Set Transformer: A Framework for Attention-based Permutation-Invariant Neural Networks}},
  booktitle = {Proc. International Conference on Machine Learning (ICML)},
  address = {Long Beach, CA, USA},
  month = {June},
  year = {2019},
  pages = {3744--3753}
}

@inproceedings{fttransformer,
  author = {Gorishniy, Yury and others},
  title = {{Revisiting Deep Learning Models for Tabular Data}},
  booktitle = {Proc. Advances in Neural Information Processing Systems (NeurIPS)},
  address = {Virtual conference},
  month = {December},
  year = {2021},
  pages = {18932--18943}
}

@inproceedings{adamw,
  author = {Ilya Loshchilov and Frank Hutter},
  title = {{Decoupled Weight Decay Regularization}},
  booktitle = {Proc. International Conference on Learning Representations (ICLR)},
  address = {New Orleans, LA, USA},
  month = {May},
  year = {2019}
}

@inproceedings{ensemble,
  author = {Lakshminarayanan, Balaji and others},
  title = {{Simple and Scalable Predictive Uncertainty Estimation using Deep Ensembles}},
  booktitle = {Proc. Advances in Neural Information Processing Systems (NIPS)},
  address = {Long Beach, CA, USA},
  month = {December},
  year = {2017},
  pages = {6402--6413}
}

@article{kim2014beam,
  author = {Kim, Joongheon and Molisch, Andreas F.},
  title = {{Fast millimeter-wave beam training with receive beamforming}},
  journal = {Journal of Communications and Networks},
  volume = {16},
  number = {5},
  pages = {512--522},
  month = {October},
  year = {2014}
}

@article{kim2015beam,
  author = {Kim, Joongheon and Hong, Song-Nam},
  title = {{Dynamic two-stage beam training for energy-efficient millimeter-wave 5G cellular systems}},
  journal = {Telecommunication Systems},
  volume = {59},
  number = {1},
  pages = {111--122},
  month = {May},
  year = {2014}
}

@article{kim2014link,
  author = {Kim, Joongheon and others},
  title = {{Fast and Low-Power Link Setup for IEEE 802.15.3c Multi-Gigabit/s Wireless Sensor Networks}},
  journal = {IEEE Communications Letters},
  volume = {18},
  number = {3},
  pages = {455--458},
  month = {March},
  year = {2014}
}

@article{han2020sharing,
  author = {Han, Sungmin and others},
  title = {{Numerical approximation of millimeter-wave frequency sharing between cellular systems and fixed service systems}},
  journal = {Journal of Communications and Networks},
  volume = {22},
  number = {1},
  pages = {37--45},
  month = {February},
  year = {2020}
}

@article{lim2024noma,
  author = {Lim, Byungju and others},
  title = {{Joint User Clustering, Beamforming, and Power Allocation for mmWave-NOMA With Imperfect SIC}},
  journal = {IEEE Transactions on Wireless Communications},
  volume = {23},
  number = {3},
  pages = {2025--2038},
  month = {March},
  year = {2024}
}

@article{kim2021multiscale,
  author = {Kim, Dohyun and others},
  title = {{Multiscale LSTM-Based Deep Learning for Very-Short-Term Photovoltaic Power Generation Forecasting in Smart City Energy Management}},
  journal = {IEEE Systems Journal},
  volume = {15},
  number = {1},
  pages = {346--354},
  month = {March},
  year = {2021}
}

@article{kim2022sar,
  author = {Kim, Kyeongrok and others},
  title = {{Stabilized Detection Accuracy Maximization Using Adaptive SAR Image Processing in LEO Networks}},
  journal = {IEEE Transactions on Vehicular Technology},
  volume = {71},
  number = {5},
  pages = {5661--5665},
  month = {May},
  year = {2022}
}

@article{yun2023detection,
  author = {Yun, Won Joon and others},
  title = {{Self-Configurable Stabilized Real-Time Detection Learning for Autonomous Driving Applications}},
  journal = {IEEE Transactions on Intelligent Transportation Systems},
  volume = {24},
  number = {1},
  pages = {885--890},
  month = {January},
  year = {2023}
}

@article{kim2023uav,
  author = {Kim, Gyu Seon and others},
  title = {{Joint frame rate adaptation and object recognition model selection for stabilized unmanned aerial vehicle surveillance}},
  journal = {ETRI Journal},
  volume = {45},
  number = {5},
  pages = {811--821},
  month = {October},
  year = {2023}
}

\end{document}